\documentclass[10pt,twocolumn,letterpaper]{article}

\usepackage[pagenumbers]{cvpr} 

\usepackage[dvipsnames]{xcolor}
\usepackage{xspace}

\usepackage{booktabs}
\usepackage{graphicx}

\usepackage{pgfplots}
\pgfplotsset{compat=1.17} 

\usepackage[flushleft]{threeparttable}

\usepackage{soulpos}
\usepackage{comment}
\usepackage{tabularx}

\DeclareMathOperator*{\mean}{\mathtt{mean}}

\newcommand{\ours}{Hier-EgoPack\xspace}

\newcommand{\ourscvpr}{EgoPack\xspace}
\newcommand{\egofourd}{Ego4D\xspace}

\usepackage{xspace}

\definecolor{up_color}{RGB}{204,51,0}
\definecolor{down_color}{RGB}{202,195,121}

\definecolor{cvprblue}{rgb}{0.21,0.49,0.74}
\usepackage[breaklinks,colorlinks,allcolors=cvprblue]{hyperref}

\title{Learning reusable concepts across different egocentric video understanding tasks}

\author{Simone Alberto Peirone
\quad
Francesca Pistilli
\quad
Antonio Alliegro
\quad
Tatiana Tommasi
\quad
Giuseppe Averta
\and
Politecnico di Torino\\
{\tt\small simone.peirone@polito.it}\\
\vspace{-10mm}
}

\begin{document}
\maketitle
\begin{abstract}
Our comprehension of video streams depicting human activities is naturally multifaceted: in just a few moments, we can grasp what is happening, identify the relevance and interactions of objects in the scene, and forecast what will happen soon, everything all at once. To endow autonomous systems with such holistic perception, learning how to correlate concepts, abstract knowledge across diverse tasks, and leverage tasks synergies when learning novel skills is essential.
In this paper, we introduce \ours, a unified framework able to create a
collection of task perspectives that can be carried across downstream tasks and used as a potential source of additional insights, as a backpack of skills that a robot can carry around and use when needed. 
Code: \href{https://github.com/sapeirone/hier-egopack}{github.com/sapeirone/hier-egopack}.
\end{abstract}    
 \vspace{-2.5mm}
\section{Introduction}\label{sec:intro}
Our daily activities are extremely complex and diverse, yet humans have the extraordinary ability to perceive, reason, and plan their actions almost entirely from visual inputs.
For instance, when observing someone at a kitchen counter with a pack of flour and a jug of water, we can infer they are kneading dough (\textit{reasoning about current activity}).
We might predict that their next step will involve mixing flour with water (\textit{reasoning about the future}) to obtain the dough (\textit{reasoning about implications}), maybe with the ultimate goal of preparing some bread (\textit{reasoning about long-range activities}).
Although natural for humans, replicating this holistic understanding in artificial intelligence remains a major challenge.
Most existing work tackles human activity understanding via task-specific models, neglecting shared reasoning patterns across tasks. Although multitask learning (MTL) offers some synergy, it struggles with negative task interference and lacks flexibility for novel tasks.
\begin{figure}[t]
    \centering
    \includegraphics[width=0.975\columnwidth,trim={0 0 0.5cm 0},clip]{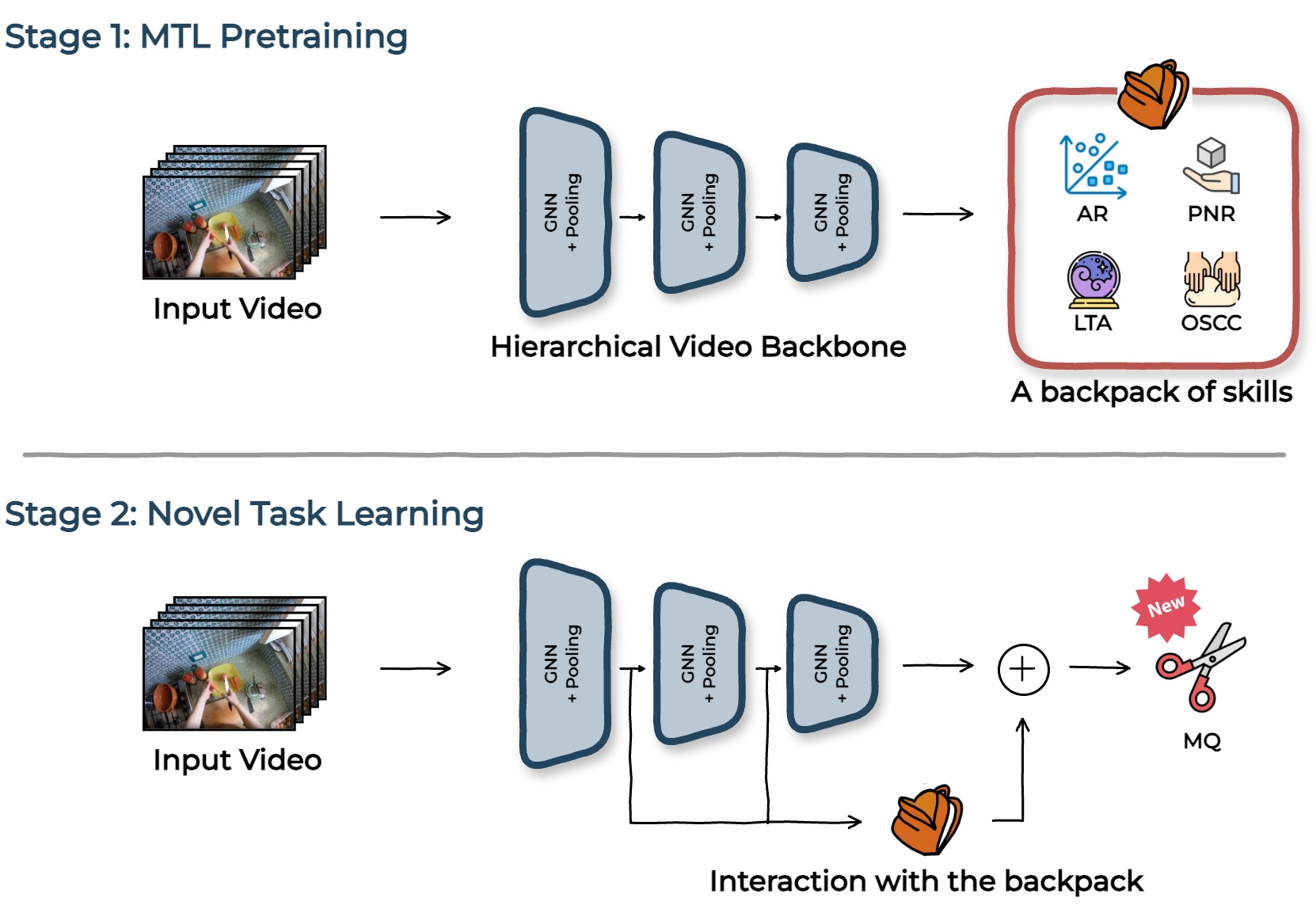}
    \caption{\textbf{Novel task learning in egocentric vision.}
    In the MTL Pretraining stage, \ours learns a set of \emph{support tasks}. Then, the knowledge from these tasks is collected in the form of prototypes and reused to foster the learning process of a \emph{novel task}.}
    \vspace{-3.05mm}
    \label{fig:teaser}
\end{figure}
We propose a paradigm shift: rather than just sharing information, systems should abstract and reuse task-specific knowledge to foster future skill learning. \ourscvpr~\cite{egopack} demonstrated this idea for egocentric videos by learning a set of reusable concepts from multiple \textit{support} tasks to enhance \textit{novel} ones. However, egocentric videos cover a wide range of tasks spanning diverse temporal scales, from actions lasting a few seconds to long-range~activities.

In this paper, we introduce \ours, an enhanced version of \ourscvpr, specifically designed to maximize positive interaction across tasks with different temporal granularity, while still using a unified architecture and minimizing task-specific weights and tuning for novel task learning (Fig.~\ref{fig:teaser}). Our hierarchical model captures both fine and coarse temporal patterns and introduces a novel Temporal Distance Gated Convolution (TDGC) layer to reason over temporal dependencies. We validate our approach on the large-scale \egofourd~\cite{ego4d} dataset, showing improved performance and positive interaction between tasks knowledge.

\section{Related works}\label{sec:related_works}

Concepts Learning covers a broad range of methods that learn an information bottleneck between the input data and the output of a desired task.
Concept Bottleneck Models (CBM)~\cite{koh2020concept} learn individual units that represent the activation of specific concepts present in the input. The concepts taxonomy may come from domain knowledge~\cite{koh2020concept}, language models~\cite{yang2023language} or obtained without any supervision~\cite{schrodi2024concept}.
Learning in the high-level concepts space may improve generalization across tasks and domains~\cite{cao2021concept} and produce more interpretable models~\cite{yang2023language}.
In video understanding, few works explored post-hoc concepts-based interpretability~\cite{kowal2024understanding} and disentanglement of static and dynamic features in action recognition models~\cite{qian2022static}. 
EgoPack~\cite{egopack} extends concepts-based learning to video understanding tasks that require different reasoning skills, collecting a set of concepts that encode how each task would ``perceive'' the same action from its specific ``perspective''.

\section{Method}\label{sec:method}

\begin{figure}[t]
    \centering
    \includegraphics[width=.925\columnwidth]{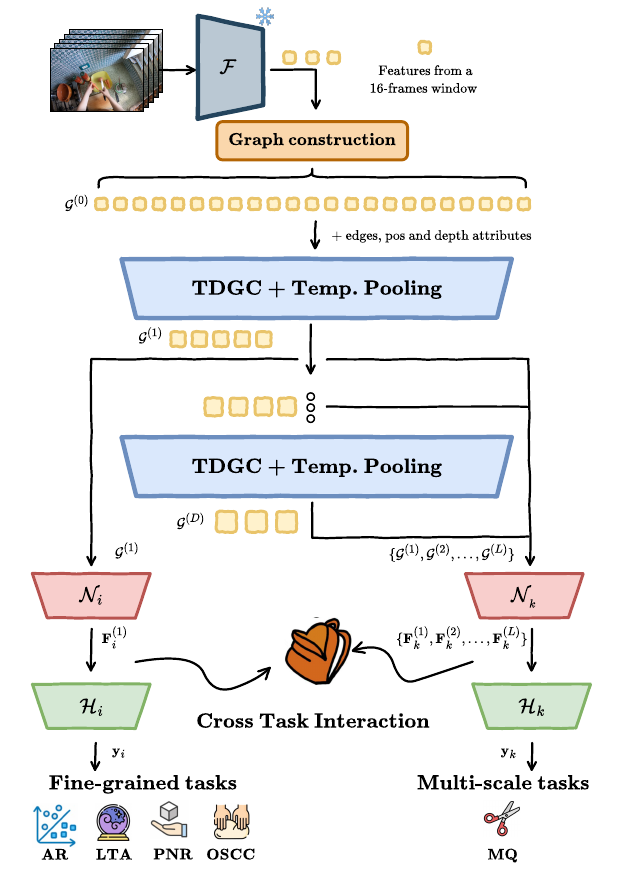}
    \vspace{-2.5mm}
    \caption{
        \textbf{Overview of \ours.}
        The video is processed as a graph by the \emph{hierarchical temporal backbone} $\mathcal{M}_t$, shared by all the tasks. 
        The node embeddings from different tasks are collected in the backpack for cross-task interaction.
    }\label{fig:architecture}
    \vspace{-2.5mm}
\end{figure}

We address a cross-task interaction setting, in which a model is trained to reuse previously acquired knowledge from a set of \emph{support tasks} to foster the learning process of any \emph{novel task}.
This work introduces \ours, a unified architecture able to model tasks with different temporal granularity and strong \emph{sense of time}, \ie reasoning on the order of events in a video.

\medskip
\noindent\textbf{Novel task learning with previous knowledge.}\label{sec:method_setting}
A task~$\mathcal{T}$ in egocentric vision is defined as a mapping between a video~$\mathcal{V}$ and an output space~$\mathcal{Y}$.
In classifications tasks, \eg, Action Recognition, this mapping assigns a trimmed video segment $v_i$ to its corresponding label~$y_i \in \mathcal{Y}$.
Differently, action localization tasks process the entire video~$\mathcal{V}$ and predict a set of temporally grounded activities, each described by its start and end timestamps and the corresponding action label: $\mathcal{T}: \mathcal{V} \to \{(t_i^s, t_i^e, y_i)\}_i.$

Our approach streamlines the processing for different tasks by feeding the temporal backbone $\mathcal{M}_t$ with untrimmed input videos and aligning the output to the downstream task at a later stage, which is a crucial design choice to enable knowledge sharing across different tasks.
We follow a two-stages training process:
\begin{itemize}
    \item \textbf{Stage 1: multi-task pretraining} on a set of $K$ support tasks to learn generalizable representations;
    \item \textbf{Stage 2: novel task learning}, in which the model adapts to a \textit{novel task} $\mathcal{T}_{K+1}$ without access to support task labels.
\end{itemize}
The key idea is to capture and reuse semantic cues shared across tasks. For example, recognizing object state changes can inform action recognition, as actions like \emph{cutting} imply change, while others like \emph{moving} may not.

\subsection{A unified architecture for Video Understanding}\label{sec:method_arch}
We represent a video $\mathcal{V}$ as a sequence of $N$ fixed-length segments with associated features $\mathbf{x} = {\mathbf{x}_1, \dots, \mathbf{x}_N}$, extracted via a pretrained video encoder $\mathcal{F}$ (e.g., EgoVLP~\cite{lin2022egocentric}).
The video can be interpreted as a temporal graph $\mathcal{G} = (\mathbf{X}, \mathcal{E}, \mathbf{pe})$, where $\mathbf{X} \in \mathbb{R}^{N \times D}$ is a matrix of features of the graph nodes, edge~$e_{ij} \in \mathcal{E}$ connects nodes~$i$ and~$j$ if their temporal distance is below $\tau$ and the attribute $\mathbf{pe} \in \mathbb{R}^{N}$ encodes the \emph{timestamp} (in seconds).
Modeling videos as graphs enables reasoning over temporal relations via message passing and to frame multiple tasks with a unified architecture.
This architecture is built on three components:
\begin{enumerate}
    \item a \emph{shared temporal backbone} $\mathcal{M}_{t}$, built with TDGC layers and subsampling for hierarchical temporal reasoning;
    \item \emph{task-specific projection necks}~$\mathcal{N}_k$ to map node embeddings to the features space of task $\mathcal{T}_k$;
    \item \emph{task-specific heads}~$\mathcal{H}_k$ for task-specific outputs.
\end{enumerate}
Let~$\mathcal{G}^{(0)}$ represent the initial graph of the input video~$\mathcal{V}$, where each node's position $\mathbf{pe}$ is initialized to the midpoint of the corresponding video segment.
Starting from $\mathcal{G}^{(0)}$, the backbone iteratively updates the graph through $L$ stages:
\begin{equation*}
    \mathcal{M}_{t}: \mathcal{G}^{(0)} \to \{\mathcal{G}^{(1)}, \mathcal{G}^{(2)}, \dots, \mathcal{G}^{(L)}\},
\end{equation*}
Each stage applies TDGC layers and temporal subsampling (mean/max pooling) to progressively enlarge the temporal extent of the nodes. Edge connections are updated based on scaled node timestamps ($\times 2^l$ at stage $l$).
The number of stages $L$ is task-dependent: single-stage for fine-grained tasks (e.g., AR, OSCC), and multi-stage for long-range temporal tasks. The architecture is shown in Fig.~\ref{fig:architecture}.

\medskip
\noindent\textbf{Temporal Distance Gated Convolution (TDGC).}
Each stage of the \emph{temporal} backbone $\mathcal{M}_{t}$ is built as a stack of~$N_l$ GNN layers, which we call Temporal Distance Gated Convolution (TDGC).
These layers are designed to preserve and encode the temporal sequence of information, capturing the relative past and future dependencies between nodes.
More specifically, given two nodes $i$ and $j$ at layer $l$, we compute $s_{ij}$ and $\mathbf{w}_{ij}$~as: 
\begin{equation*}
    s_{ij} = \mathtt{sign}(\mathbf{pe}_{[i]}^{(l)} - \mathbf{pe}_{[j]}^{(l)}), \;\;\; \mathbf{w}_{ij} = \mathtt{MLP}(|\mathbf{pe}_{[i]}^{(l)} - \mathbf{pe}_{[j]}^{(l)}|).
\end{equation*}
These two factors are used to re-weight the contribution of each neighbor node $j$ in the aggregation step, as follows:
\begin{align*}
    \mathbf{x}_j^{'}     & = \mathtt{MLP}\left(\mathbf{x}_j^{(l)}\right) = \phi(\mathbf{W}_n^T \mathbf{x}_j^{(l)} + \mathbf{b}_n),                                                  \\
    \mathbf{x}_i^{(l+1)} & = \mathbf{W}^T_r\mathbf{x}_i^{(l)} + \mean_{j \in \bar{\mathcal{N}}(i)} \left( s_{ij} ( \mathbf{w}_{ij}  \odot \mathbf{x}_j^{'}) \right) + \mathbf{b}_r,
\end{align*}
where $\mathbf{x}_i^{(l)}$ are the features of the node $i$ at layer $l$, $\bar{\mathcal{N}}(i)$ is the set of neighbors of node $i$, $\mathbf{W}_n$, $\mathbf{W}_r$ and $\mathbf{b}_n$, $\mathbf{b}_r$ are learnable weights and biases respectively. 

\subsection{Task-specific components}\label{sec:method_ts}
The temporal backbone $\mathcal{M}_t$ is shared across all downstream tasks and provides task-agnostic temporal reasoning over streams of fixed-length video segments.
Each task $\mathcal{T}_k$ has its own projection neck $\mathcal{N}_k$, a two-layer MLP that maps backbone outputs to the task's feature space: $
    \mathbf{X}^{(l)}_k = \mathcal{N}_k \left( \mathbf{X}^{(l)} \right) \;\text{with}\;\mathcal{N}_k: \mathbb{R}^D \to \mathbb{R}^D. $
For tasks with known temporal boundaries (\eg, AR), we align node embeddings to task annotations: 
\begin{equation*}
    \mathbf{F}_{k,[i]}^{(l)} = \mathtt{align} (\mathbf{X}^{(l)}_k, s_i, e_i) = \mean_{j: \;s_i<\mathbf{p}^{(l)}_{[j]}<e_i} \mathbf{X}^{(l)}_{k,[j]},
\end{equation*}
where $\mathbf{F}_{k,[i]}^{(l)}$ are the task-specific features of segment $v_i$ for task $\mathcal{T}_k$.
For action localization, which operate over the full video, alignment is unnecessary and we directly use $\mathbf{X}^{(l)}_k$.

To solve the \emph{novel task} $\mathcal{T}_{K+1}$, the naive approach would be to finetune the model, adding new task-specific neck $\mathcal{N}_{K+1}$ and head $\mathcal{H}_{K+1}$ and possibly updating the temporal backbone $\mathcal{M}_{t}$.
However, this approach may forget previously acquired knowledge.
Instead, we explicitly model the perspectives of the \emph{support tasks}, learned during the MTL pre-training step, as a set of task-specific prototypes that can be accessed by the novel task. 
We collect these task-specific prototypes from videos annotated for action recognition, as human actions can be seen as the common thread behind the different tasks. We forward these action samples through the temporal backbone, align them based on the AR annotations and project their features using the task-specific necks $\mathcal{N}_k$ of each task to obtain the task-specific features $\mathbf{F}_k$, each row capturing that task’s “perspective” on a given segment.
We then aggregate features by action label (a unique verb-noun pair) to form the prototypes set
    $\mathbf{P}^k = \{ \mathbf{p}^{k}_{0}, \mathbf{p}^{k}_{1}, \dots, \mathbf{p}^{k}_{P
    } \} \in \mathbb{R}^{P \times D}$
for each task $\mathcal{T}_k$, where $P$ is the number of unique verb-noun pairs and $D$ is the feature dimension.
These prototypes are frozen and serve as a compact summary of each task’s learned representation—a reusable knowledge base for novel tasks, like a \emph{backpack of skills} that the model can carry over.

\begin{figure*}[htbp]
    \hfill
    \begin{minipage}{0.235\textwidth}
        \includegraphics[trim=0.5cm 0.5cm 0.5cm 0.5cm,height=3.5cm]{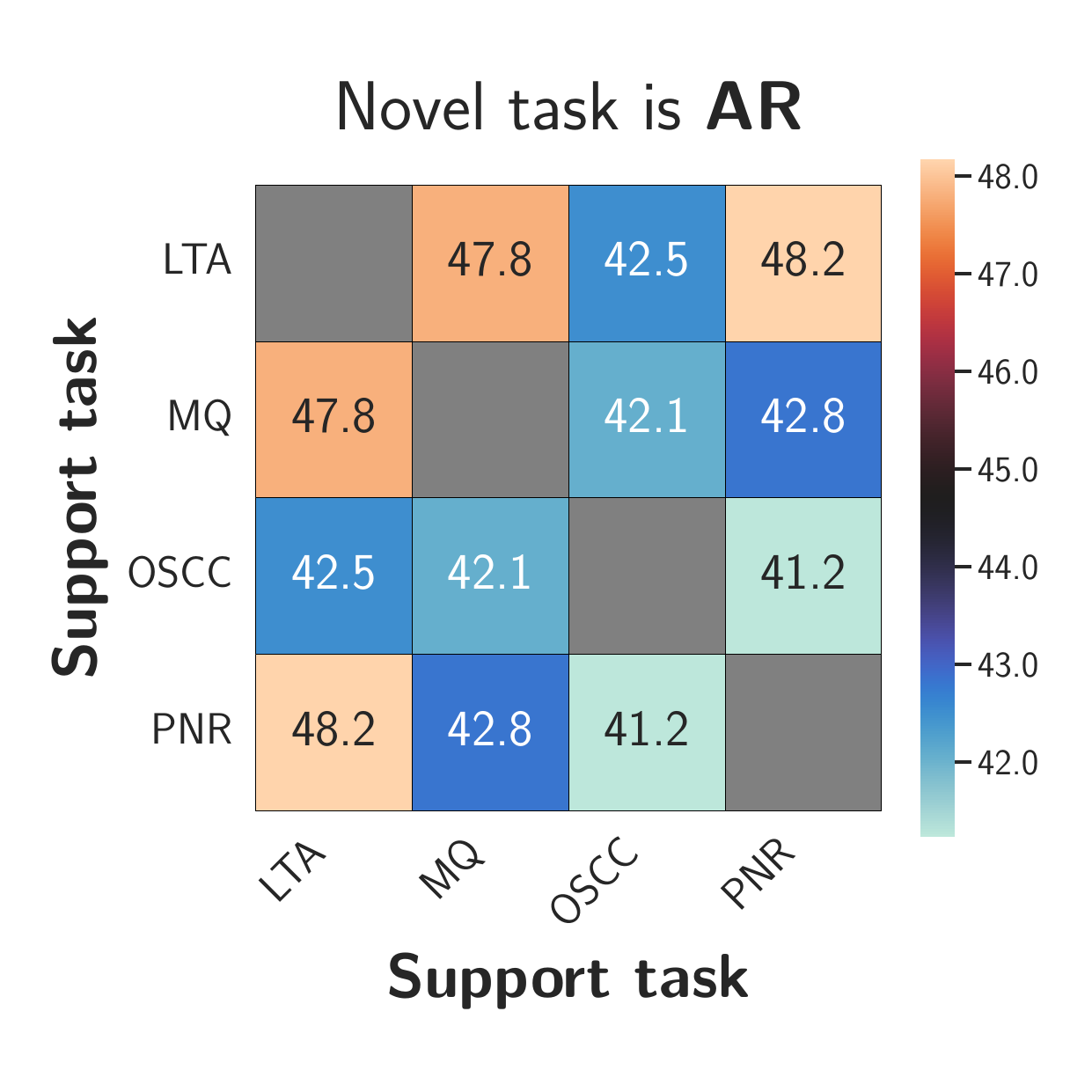}
    \end{minipage}
    \hfill
    \begin{minipage}{0.235\textwidth}
        \includegraphics[trim=0.5cm 0.5cm 0.5cm 0.5cm,height=3.5cm]{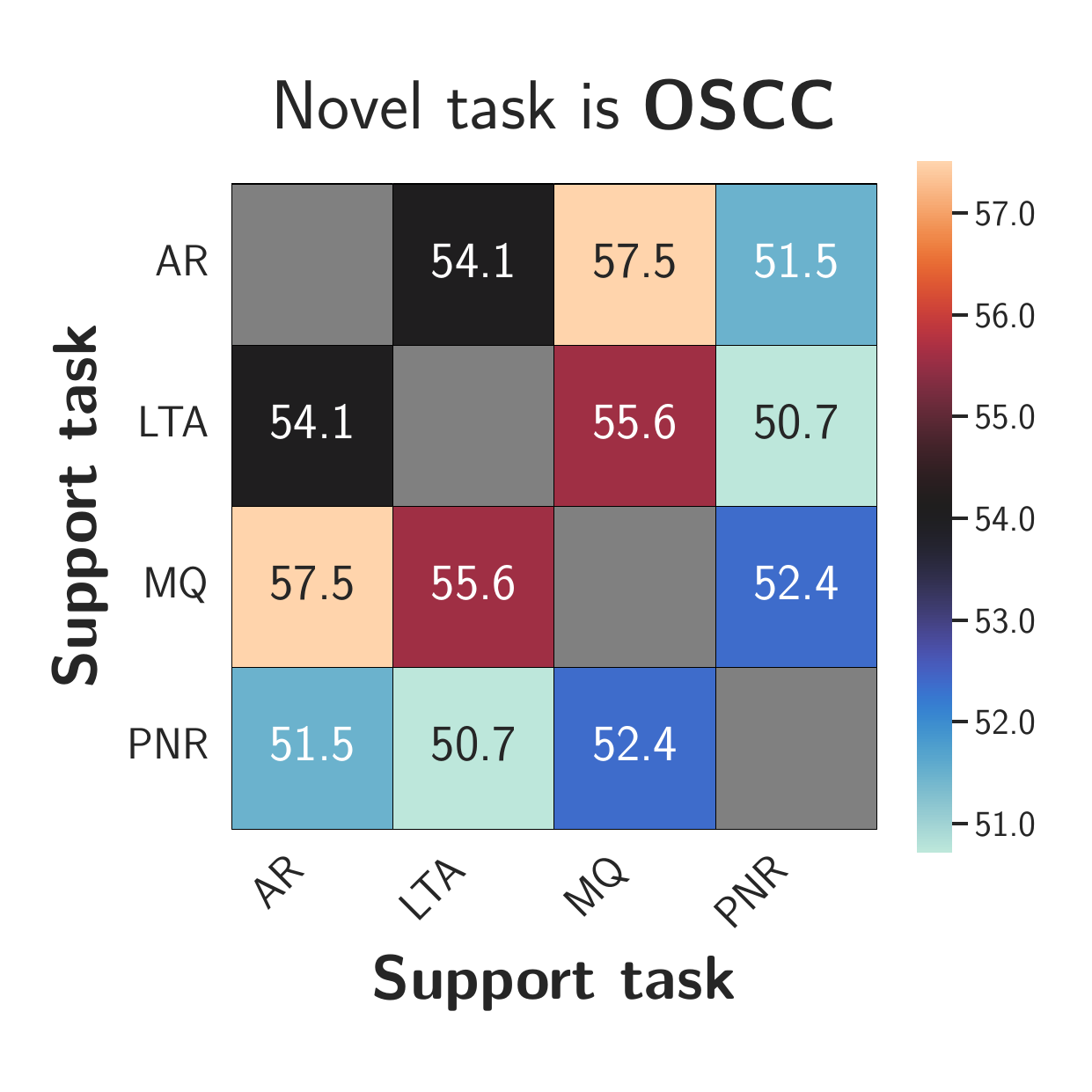}
    \end{minipage}
    \hfill
    \begin{minipage}{0.235\textwidth}
        \includegraphics[trim=0.5cm 0.5cm 0.5cm 0.5cm,height=3.5cm]{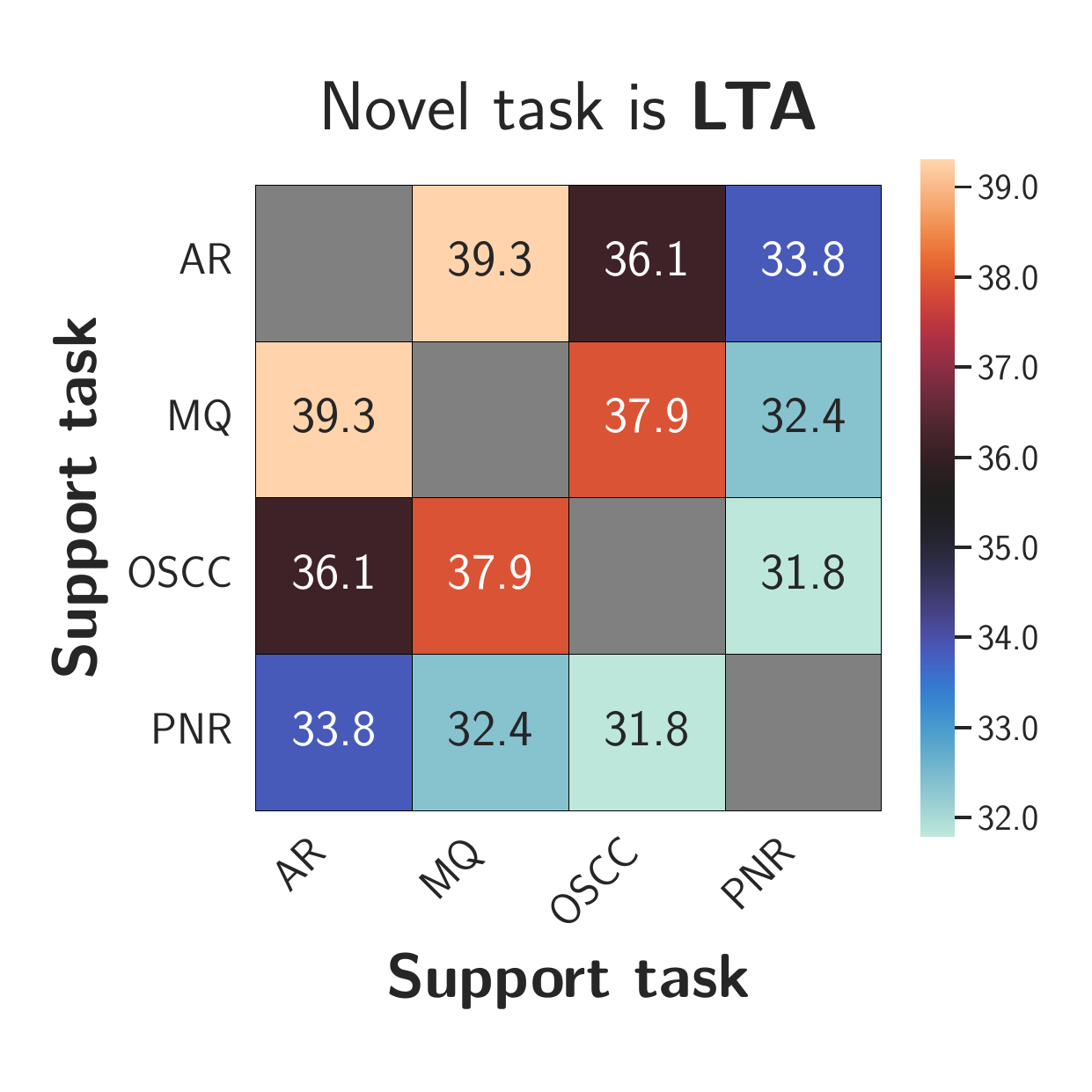}
    \end{minipage}
    \hfill
    \begin{minipage}{0.235\textwidth}
        \includegraphics[trim=0.5cm 0.5cm 0cm 0.5cm,height=3.5cm]{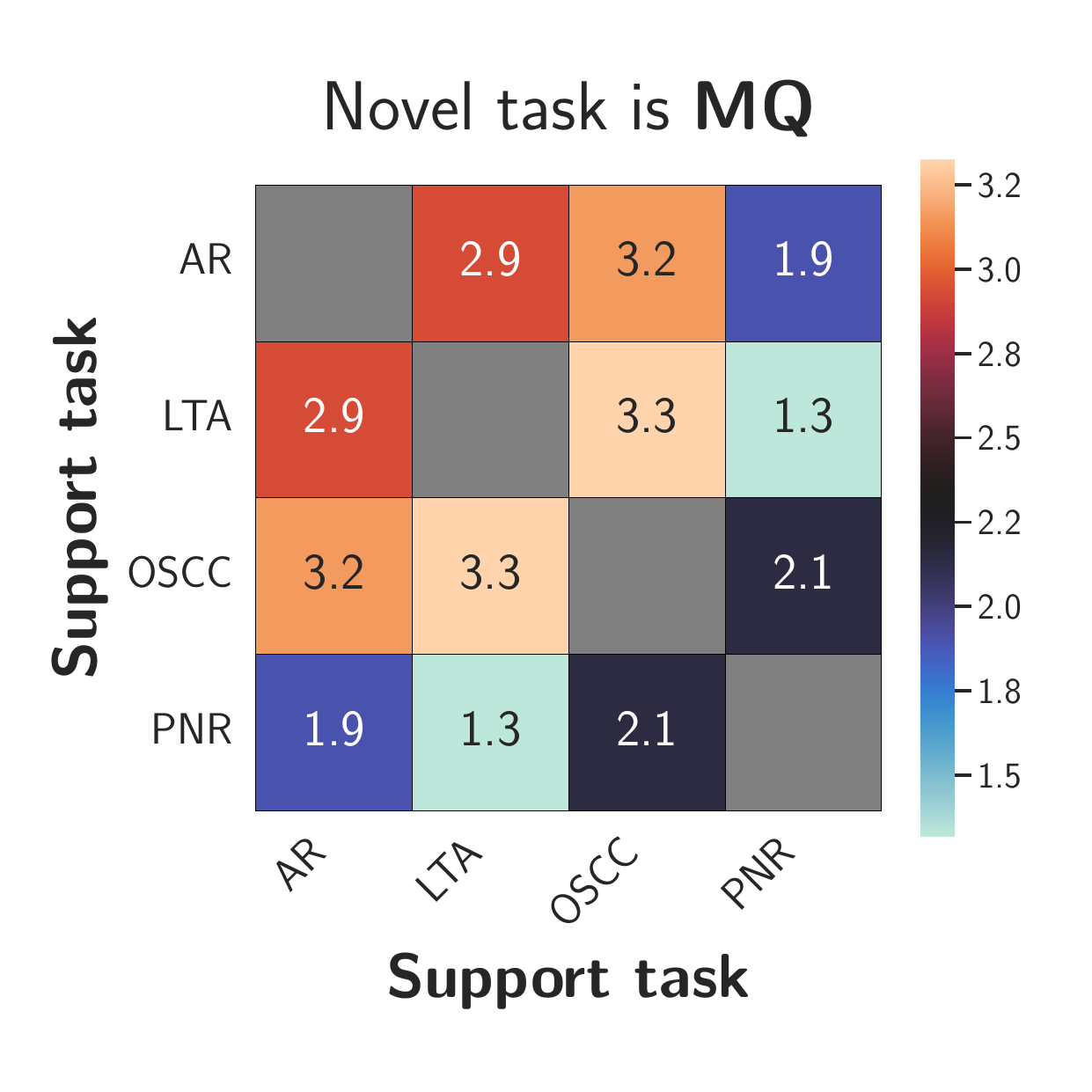}
    \end{minipage}
    \hfill
    \vspace{-3.5mm}
    \caption{\textbf{Activations consensus for different \emph{novel tasks}.} Activations consensus between two \emph{support tasks} is defined as the percentage of their prototypes corresponding to the same label activated by the two tasks. 
    }
    \label{fig:consensus}
\end{figure*}

\subsection{Learning a novel task with a backpack}\label{sec:method_egopack_learning}
To solve a novel task $\mathcal{T}_{K+1}$, we pass the output graphs from the temporal backbone through all the task-specific necks
to obtain features $\mathbf{X}_k$.
These features act as queries to retrieve the closest task prototypes in $\mathbf{P}^k$ via k-NN search in the feature space. Each query and its neighboring prototypes form a graph-like structure, where message passing is applied using $M$ layers of SAGE convolution to iteratively refine the task-specific features.
At each layer $m$, we update the features~$\mathbf{X}_{k,[i]}^{(m)}$: 
\begin{equation*}
    \mathbf{X}^{(m+1)}_{k,[i]} = \mathbf{W}^{(m)}_{r} \mathbf{X}^{(m)}_{k,[i]} + \mathbf{W}^{(m)} \cdot  \mean_{\mathbf{p}^k_{j} \, \in \, \bar{\mathcal{N}}(i)} \mathbf{p}^k_{j},
\end{equation*}\label{eq:egopackgnn}
where~$\mathbf{p}^k_{j} \, \in \, \bar{\mathcal{N}}(i)$ is the set of \emph{activated prototypes} for the given task, and $\mathbf{W}^{(m)}_{r},\mathbf{W}^{(m)}$ are learnable projections. This refinement is applied at each backbone stage $l$, updating only the task features—not the prototypes—to preserve the original learned perspectives. The final refined features $\tilde{\mathbf{X}}^{(l)}_k$ are aligned, if needed, to produce $\tilde{\mathbf{F}}^{(l)}_k$.
We evaluate different fusion strategies to integrate the novel task with the perspectives gained from the previous tasks.
In \emph{features-level} fusion, we average the task-specific features for the novel task $\mathbf{F}_{K+1}$ with the \textit{refined} perspectives from the previous tasks~$\tilde{\mathbf{F}}_k$.
In \emph{logits-level} fusion, we keep a set of separate heads, one for each support task, feed the features $\tilde{\mathbf{F}}_k$ to each head separately and sum their outputs. 
Intuitively, this approach allows each task to cast a vote on the final prediction, based on its perspective on the same video segment.

\section{Experiments}\label{sec:experiments}

We validate \ours on \egofourd~\cite{ego4d}, focusing on five benchmarks that cover different temporal granularities.
\emph{Fine-grained tasks} focus on short-term understanding of the video, usually a few seconds long, and include: \emph{Action Recognition (AR)}, \emph{Object State Change Classification (OSCC)}, \emph{Point of No Return (PNR)}, \emph{Long Term Anticipation (LTA)}.
Other tasks may require both short and long term understanding of the input video.
Among these, we analyze an action localization task, \ie, \emph{Moment Queries (MQ)}.

\begin{table}[tb]
  \centering
  \footnotesize
  \caption{\ours on a set of \egofourd tasks.}
  \label{tab:main_results}
  \vspace{-0.25cm}
  \resizebox{0.49\textwidth}{!}{
  \begin{tabular}{lccccccc}
    \toprule
    & \multicolumn{2}{c}{\textbf{AR Top-1 (\%)}} & \textbf{OSCC Acc.} & \multicolumn{2}{c}{\textbf{LTA ED} ($\downarrow$)} & \textbf{PNR Err.} & \textbf{MQ mAP} \\
    \cmidrule(lr){2-3} \cmidrule(lr){4-4} \cmidrule(lr){5-6} \cmidrule(lr){7-7} \cmidrule(lr){8-8}
    \textbf{Method} & Verb & Noun & (\%) & Verb & Noun & ($\downarrow$) & (\%) \\
    \midrule
    Ego4D Baselines~\cite{ego4d}     & 22.18 & 21.55 & 68.22 & 0.746 & 0.789 & \underline{0.62} & 6.03 \\
    EgoT2s~\cite{egot2}              & 23.04 & 23.28 & 72.69 & 0.731 & 0.769 & \textbf{0.61}    & N/A \\
    \ourscvpr~\cite{egopack}         & 25.10 & 31.10 & 71.83 & \underline{0.728} & \underline{0.752} & \textbf{0.61}    & N/A \\
    \midrule
    \midrule
    Single Task                      & \underline{26.93} & 33.50 & \underline{75.22} & \underline{0.728} & \underline{0.752} & \underline{0.62} & \underline{20.2} \\
    MTL                              & 26.31 & \underline{33.90} & 74.79 & 0.730 & 0.754 & \underline{0.62} & 18.5 \\
    \midrule
    \textbf{\ours}                   & \textbf{27.30} & \textbf{34.65} & \textbf{75.60} & \textbf{0.725} & \textbf{0.741} & \textbf{0.61} & \textbf{21.0} \\
    \bottomrule
  \end{tabular}}
  \begin{tablenotes}
    \scriptsize
    \item \emph{Single Task} uses the same hierarchical GNN-based architecture to model all tasks. \emph{Multi-Task Learning (MTL)} uses hard parameter sharing to jointly learn all tasks.
  \end{tablenotes}
  \vspace{-3mm}
\end{table}

\medskip
\noindent\textbf{Quantitative results.}
We show the main results in Table~\ref{tab:main_results}, comparing our approach with the \egofourd baselines~\cite{ego4d}, the task-translation framework EgoT2~\cite{egot2} and \ourscvpr~\cite{egopack}.

We observe that the task prototypes in \ours provide a comprehensive and easy-to-access abstraction of the model’s learned knowledge, enabling the extraction of relevant insights tailored to the specific sample and task, exhibiting superior performance.

\smallskip
\smallskip

\medskip
\noindent\textbf{Activation consensus across tasks.}
We analyze how \ours leverages knowledge abstractions from the \emph{support tasks}. 
Specifically, we visualize the \emph{activated prototypes} (\ie the set of prototypes each \emph{support task} looks at) during the interaction process of \ours across different novel tasks and quantify task \emph{activation consensus}, \ie, the degree to which different tasks activate prototypes corresponding to the same label for a given sample of the \emph{novel task}.
A low consensus suggests that the support tasks capture more diverse cues, \ie different tasks activate different prototypes, whereas a high consensus indicates that activations are more coherent across tasks.
Fine-grained tasks, \eg, AR, have higher average consensus compared to MQ (Fig.~\ref{fig:consensus}).
We attribute this difference to the implementation of the interaction process for these two groups of tasks. 
In fine-grained tasks, the interaction process is applied on the sample-level aligned features, while we use node-level features in MQ which may correspond to background or poorly discriminating regions of the video.
However, the low average activations consensus and high diversity in prototypes' activations across tasks
shows how \ours is effectively integrating different perspectives for the MQ~task.

\smallskip
\smallskip

\section{Conclusions}\label{sec:conclusions}
We present \ours, an holistic video understanding model that enables knowledge sharing between egocentric vision tasks with different temporal granularity.
Our work emphasizes the importance of prior knowledge and task perspectives in learning novel tasks, focusing on how task-specific knowledge is represented and utilized. Moreover, through our proposed unified architecture, we demonstrate that leveraging diverse task perspectives in egocentric vision, even across varying temporal granularity, leads to more comprehensive and human-like video understanding.

\section*{Acknowledgments}
This study was carried out within the FAIR - Future Artificial Intelligence Research and received funding from the European Union Next-GenerationEU (PIANO NAZIONALE DI RIPRESA E RESILIENZA (PNRR) – MISSIONE 4 COMPONENTE 2, INVESTIMENTO 1.3 – D.D. 1555 11/10/2022, PE00000013). This manuscript reflects only the authors’ views and opinions, neither the European Union nor the European Commission can be considered responsible for them. 
Antonio Alliegro and Tatiana Tommasi also acknowledge the EU project ELSA - European Lighthouse on Secure and Safe AI (grant number 101070617).

{
    \small
    \bibliographystyle{ieeenat_fullname}
    \bibliography{main}
}


\end{document}